\documentclass{Interspeech}
\graphicspath{ {./figures/} }

\usepackage{xspace}
\usepackage{multirow}
\newcommand{\method}{\textsc{WhiStress}\xspace}
\newcommand{\dataset}{\textsc{TinyStress-15K}\xspace}
\newcommand{\newpara}[1]{\vspace{0.07cm}\noindent \textbf{#1}}

\usepackage{acronym}

\acrodef{ASR}{Automatic Speech Recognition}
\acrodef{TTS}{Text-to-Speech}
\acrodef{LLM}{Large Language Model}
\acrodef{LM}{Language Model}

\renewcommand{\thefootnote}{}%

\interspeechcameraready 

\title{\method: Enriching Transcriptions with Sentence Stress Detection}

\author{*Iddo}{Yosha}
\author{*Dorin}{Shteyman}
\author{Yossi}{Adi}

\affiliation{The School of Computer Science and Engineering}{}{}
\affiliation{The Hebrew University of Jerusalem, Israel}{}{}

\email{\{iddo.yosha, dorin.shteyman\}@mail.huji.ac.il}


\usepackage{comment}

\begin{document}

\maketitle

\begin{abstract}
Spoken language conveys meaning not only through words but also through intonation, emotion, and emphasis. Sentence stress, the emphasis placed on specific words within a sentence, is crucial for conveying speaker intent and has been extensively studied in linguistics. In this work, we introduce \method, an alignment-free approach for enhancing transcription systems with sentence stress detection. To support this task, we propose \dataset, a scalable, synthetic training data for the task of sentence stress detection which resulted from a fully automated dataset creation process. We train \method on \dataset and evaluate it against several competitive baselines. Our results show that \method outperforms existing methods while requiring no additional input priors during training or inference. Notably, despite being trained on synthetic data, \method demonstrates strong zero-shot generalization across diverse benchmarks. Project page: \url{https://pages.cs.huji.ac.il/adiyoss-lab/whistress}. \begingroup
\renewcommand\thefootnote{}\footnote{*Equal contribution.}%
\addtocounter{footnote}{-1}
\endgroup
\end{abstract} 

\section{Introduction}
\label{sec:intro}
Theoretical work on sentence stress can be divided into linguistic and acoustic research. The linguistic formulation of stress falls into two perspectives, as described in \cite{Ladd_2008}. The first, defines normal stress as a default pattern, independent of meaning, that follows phonological constraints \cite{Chomsky1968TheSP}. The second perspective views sentence stress as a semantic tool, where stress may be placed on any word to highlight its semantic importance \cite{Bolinger}. Furthermore, acoustic research has established that sentence stress is manifested in the speech signal mainly through variations in duration, amplitude and pitch \cite{AcousticCorrelates}.

\begin{figure}[t!]
    \centering
    \includegraphics[width=0.49\textwidth]{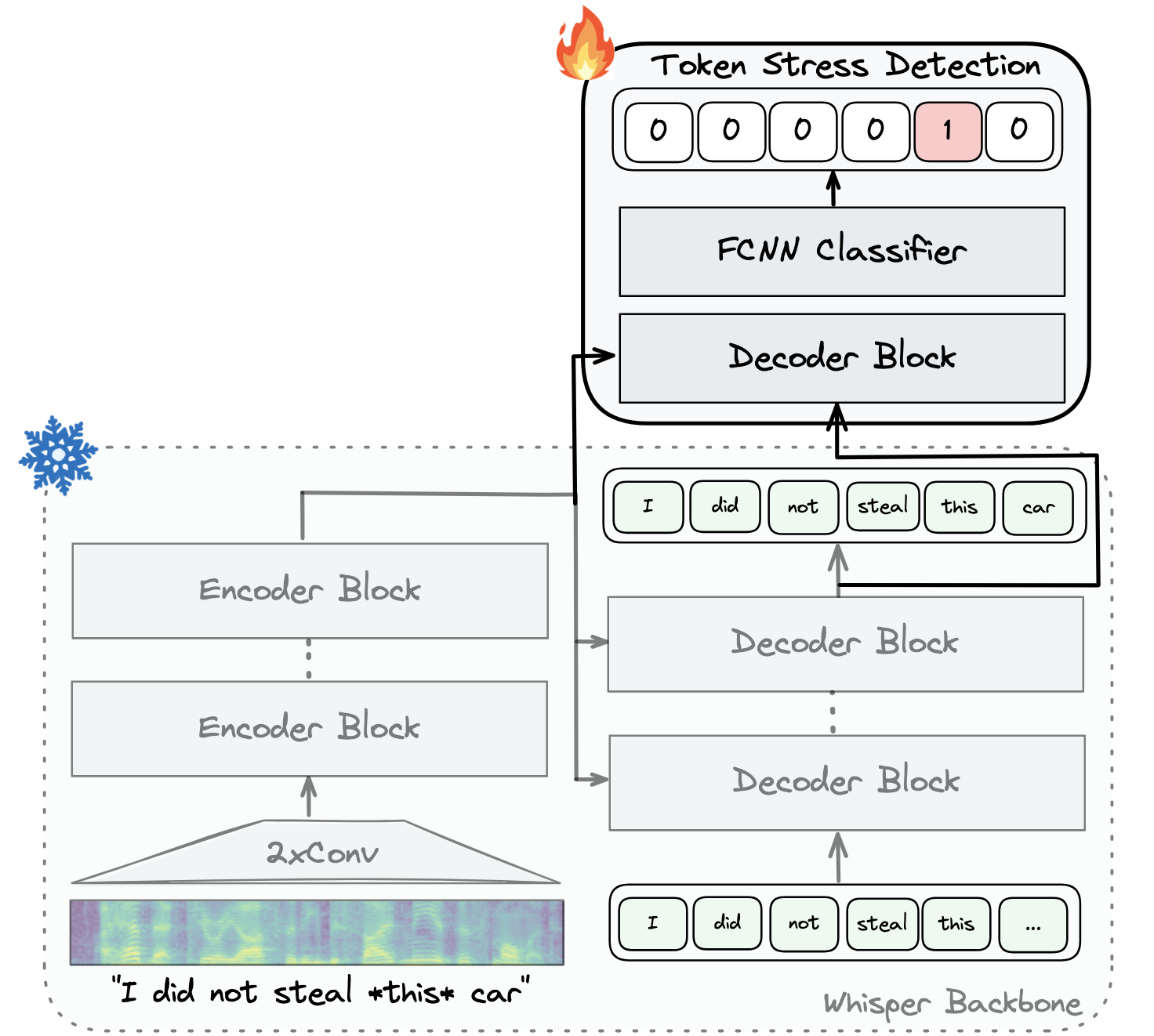}
    \caption{\method Architecture. The Whisper model is kept frozen during training. The extension is a  transformer decoder block with cross-attention 
    for the audio embeddings, followed by an FCNN classifier that outputs the stress score per token.\label{fig:model_architecture}}    
\end{figure}

Despite these linguistic insights, some computational models for sentence stress detection have relied on acoustic features, with limited integration of linguistic information while using both classical machine learning methods~\cite{ProsodicStressRevisited, 3PROarticle, mishra12_interspeech} and deep learning methods~\cite{morrison2023crowdsourcedautomaticspeechprominence, deseyssel2024emphassessprosodicbenchmark}. Other works show that leveraging language priors, such as phrase and word boundaries, intonation units, and syntactic features can improve sentence stress prediction~\cite{Lin2020JointDO, koreanarticle, suni2015hierarchical}.
Although demonstrating impressive sentence stress prediction performance, the above-mentioned models require during inference one or more of the following: (i) transcriptions of the spoken utterances, (ii) word boundaries, via forced alignment or manual annotation. This reliance introduces a critical constraint, as model performance is influenced by the accuracy of the forced aligner, transcription quality and data scalability. 
As an exception to these, the authors of \cite{biron2024non}, suggested fine-tuning Whisper \cite{radford2022robustspeechrecognitionlargescale} by adding tokens in the text for stress labeling. However, their method do not take into account possible performance degradation caused by altering transcriptions and the fine-tuning process \cite{mai2024finetuningfinecalibrated}. Another notable work, attempted to use BERT \cite{devlin2019bertpretrainingdeepbidirectional} to predict stress at the token level directly from text for controlled prominence in \ac{TTS} \cite{stephenson2022bertpredictcontrastivefocus}.

In this work, we introduce \method, a novel, alignment-free approach for sentence stress detection. The proposed approach leverages a \ac{LM} conditioned on acoustic signals to improve sentence stress detection. We focus on English speech, and equip the Whisper model~\cite{radford2022robustspeechrecognitionlargescale} with an additional stress detection head, that predicts stress targets for each token, thus allowing the model to generate more informative transcriptions without affecting the original model's performance. 
See Figure~\ref{fig:model_architecture} for a visual description. 

Another significant factor in determining the performance, generalization, and robustness of any proposed model is the quality of training data. Most of the current models that detect sentence stress were trained on either: (i) closed source data \cite{deseyssel2024emphassessprosodicbenchmark}; (ii) datasets with non-standard emphasis annotations \cite{Lin2020JointDO,koreanarticle}; or (iii) crowdsourced relying on human annotator judgment to annotate stressed words \cite{morrison2023crowdsourcedautomaticspeechprominence}. Consequently, the quality and consistency of the data can potentially lead to suboptimal performance of any model trained for sentence stress detection. 

This motivated the development of \dataset, a scalable, synthetically generated dataset, designed for sentence stress detection amounting to $\sim$15 hours of speech. \dataset was derived from text data, with sentence-stress annotations carrying semantic and rhetorical meaning, generated by a large language model. The corresponding speech was then synthesized using a text-to-speech service, with altered prosodic features to replicate natural vocal stress. This automated approach enabled the generation of a large, diverse dataset, tailored for training stress detection models.

\newpara{Our contributions:} (i) We present \method, an alignment-free approach for token-level stress classification in \ac{ASR}, which can be extended to other paralinguistic and language tasks that predict targets at the word level in future research; (ii) We introduce a method for generating synthetic, stress-annotated speech data to advance future research; (iii) We provide valuable analysis of the Whisper backbone layers to highlight a potential prosodic-linguistic trade-off for sentence stress detection in \method; and (iv) We publicly release our code, \method model weights and \dataset dataset.

\section{Synthetic Data}
\label{sec:data}

We develop an automated pipeline for synthetic dataset generation, consisting of three key steps: (i) transcription selection, (ii) stressed words labeling, and (iii) speech synthesis. 

\newpara{Transcription Selection.} To get transcriptions of coherent, diverse, everyday language, we use the TinyStories dataset, a common choice for small language models training \cite{eldan2023tinystoriessmalllanguagemodels}. Each sample in the tinyStories dataset is a paragraph of a short story. We generate each sample as a sentence extracted from each paragraph using the NLTK library \cite{10.5555/1717171}. Sentences with fewer than three words are filtered out.

\newpara{Stressed Words Labeling.}
We chose which words to emphasize by instructing GPT-4o-mini~\cite{hurst2024gpt} to provide two different options for stressed words in each sentence, such that selected words reflect natural sentence stress, meaning they influence the sentence's interpretation in a semantically significant way.
In the training set, we used both stress patterns, while in the test set we randomly chose one stress pattern of the two.

\newpara{Speech Synthesis.} We use the Google Text-to-Speech API to synthesize the stressed speech based on the prior steps, by using SSML syntax that enables editing prosodic features on the word or constituent level.\footnote{SSML (Speech Synthesis Markup Language) is a standardized markup language used to control various aspects of speech synthesis for \ac{TTS} systems.}
The following steps were taken: (i) \emph{Prosodic features adjustment:} A core step in the data creation procedure. Every emphasized word in the transcription has its amplitude, duration and pitch adjusted. Initially, the adjustments were based on empirical results from \cite{AcousticCorrelates}. However, since the TTS service generated overly robotic speech, we manually refine them using the following heuristics to achieve a more natural sound: The volume and duration are in proportion  to the length of the word, with shorter words having a higher volume and longer duration than longer words. Specifically, duration changed by reducing the speaking rate by $30\%$ to $85\%$ relative to the original rate, and the volume was increased by $3$ to $6$db. The pitch of emphasized words was adjusted to be higher by $1.5$ semitones. To introduce variability, all prosodic features were augmented by adding random standard Gaussian noise.; (ii) \emph{Speaker voice selection:} To ensure diversity and fairness in the dataset, each sample was synthesized by a randomly chosen female or male speaker from a pool of $10$ possible voices; and (iii) \emph{Timestamps annotations:} We generate start timestamps for each word in the transcription using the \ac{TTS} service, to get a word level time alignment.

Overall, we result in a dataset of $1$k test samples and $15$k training samples which we denote as \dataset. The total time of synthesized speech in the training set is $\sim$15 hours.
\section{Method}
\label{sec:method}

\method enhances transcription models with additional sentence stress detection objective. Specifically, we propose generating two outputs for each input speech signal: the transcription and an emphasis score for each token in the transcription. To achieve this, we modify both the architecture of the Whisper model and its training procedure.

\subsection{Architecture}
\label{sec:Architecture}
We propose a modification to the Whisper model \cite{radford2022robustspeechrecognitionlargescale} by incorporating a stress detection head that learns to identify sentence stress in the speech signal for each transcribed token. As shown in Figure \ref{fig:model_architecture},
\method model consists of two components:

\newpara{Whisper Model.} The backbone of the \method model. It is used to process raw audio into hidden representations that encode phonetic, linguistic, and prosodic features. These representations serve a dual purpose: (i) input for the stress detection head; and (ii) producing Whisper's speech transcription.

\newpara{Stress Detection Head.} A learnable component extending the Whisper model, consisting of: (i) Whisper Decoder Block; and (ii) Fully Connected Neural Network (FCNN) Classifier. The additional Whisper decoder block applies cross-attention between decoder and encoder hidden states of the backbone Whisper model. This component learns the acoustic and linguistic features that contribute to the stress detection. According to our findings in Section \ref{sec:analysis}, the inputs to the additional decoder head are the encoder and decoder embeddings of the $9$th layer of the backbone Whisper model. The FCNN classifier is two-layer fully connected neural network that acts as a binary classifier, processing the output of the additional decoder block to assign a stress label to each token ($1$ if stressed, $0$ otherwise).

\subsection{Training}
\label{sec:Alignment}
\newpara{Label Alignment.} As a first step, the ground-truth word-level stress labels are converted to token-level labels, aligned with error-free transcription tokens. However, Whisper-generated hidden states, which may encode transcription errors, serve as input to the stress detection head. Therefore, any Whisper transcription error can propagate and misalign stress labels by shifting decoder input tokens. Meanwhile, samples with minor transcription errors at word-level remain acoustically informative. To retain such samples, we filter out training examples where the Whisper-generated transcription word length differs from the ground truth word length, allowing for word-level transcription errors. The described length filtering approach mitigates mislabeling stressed words thus ensures reliable supervision.

Unlike previous methods, our label-alignment procedure relies solely on a relaxed word-to-word matching between ground truth (i.e., stress labels) and generated transcriptions as a pre-processing step before training, and is alignment-free during inference. Notably, at any stage, it does not require word-level timestamps (i.e., no time alignment is needed), as Whisper inherently aligns generated tokens with audio features.

\newpara{Training Procedure.} During training, the backbone Whisper model remains frozen, and only the stress detection head (Section~\ref{sec:Architecture}) is trained using cross-entropy loss. The number of trainable parameters in the additional head is $\sim7$ million. Additionally, non-word tokens, such as punctuation marks, are included to help \method identify them as unstressed.
\section{Model Analysis}
\label{sec:analysis}

We analyze Whisper's internal representations to determine which layers capture prosodic features, specifically pitch, energy, and duration. Furthermore, we investigate how \method identifies sentence stress by evaluating different layers as input for the stress detection head.

\subsection{Prosodic information analysis}
\begin{figure}
    \centering
    \includegraphics[width=0.4\textwidth]{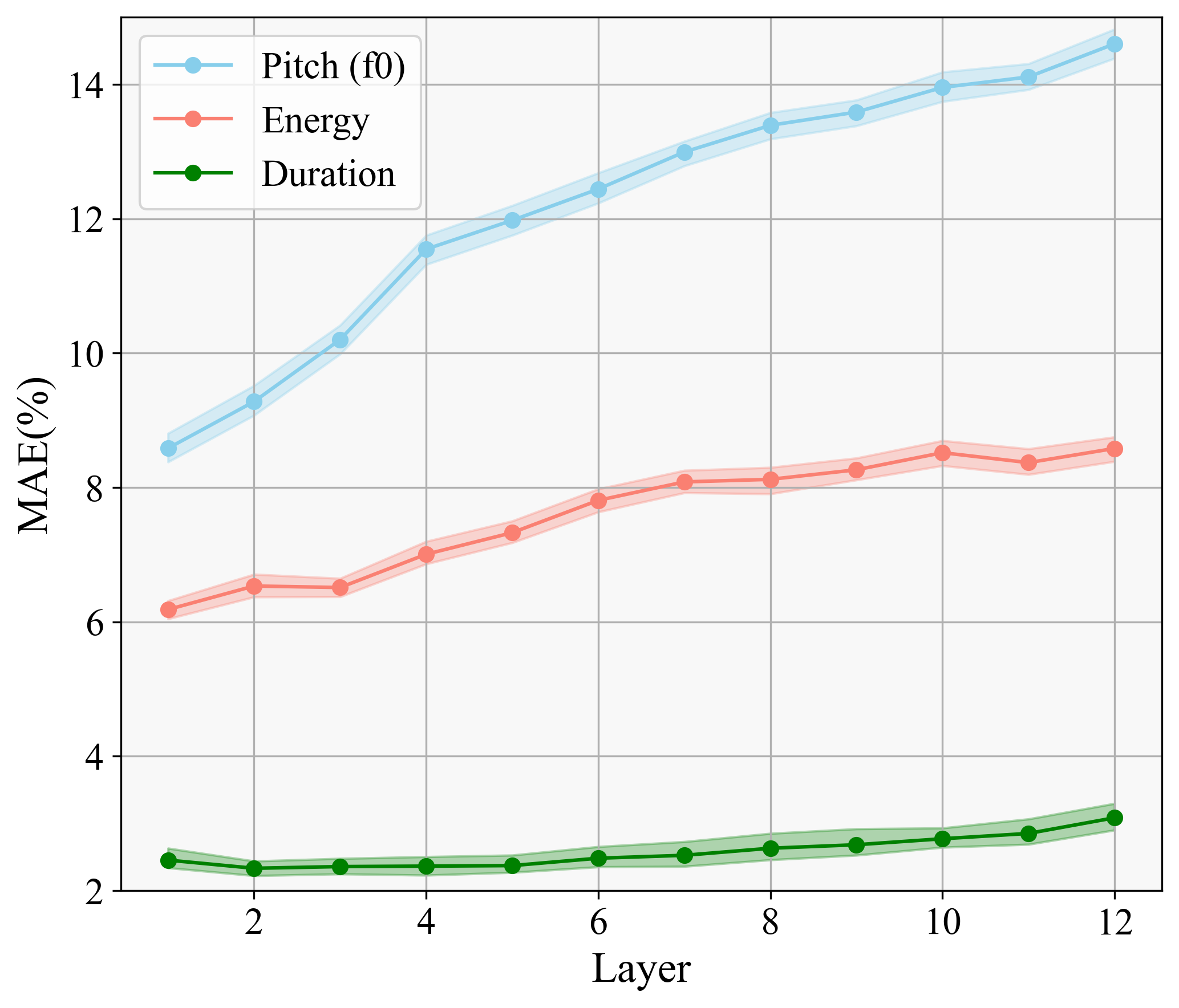}
    \caption{
    Prosodic features prediction by Mean Absolute Error (MAE) percentage of Whisper layer embeddings. A lower MAE percentage indicates better prediction. Each curve shows confidence intervals.
    \label{fig:encoder_decoder_analysis}
    }
    \vspace{-0.2cm}
\end{figure}

To understand where prosodic information is stored in \method, we analyze the embeddings of the Whisper backbone model. We use a subset of CREMA-D dataset \cite{6849440} that contains speech samples with varying emotional content, making prosodic features more prominent in the signal.

To analyze the energy and pitch targets, we utilize the Whisper encoder embeddings, which capture pure acoustic features without conditioning on transcription. We compute fundamental frequency (F$0$) and root mean square (RMS) energy in $75$ ms windows with a $20$ ms stride to align with the frame rate of the audio embeddings. To construct targets, we apply max pooling to F$0$ and mean pooling to RMS energy over $300$ ms windows. For each window, at every layer, we pool the mean encoder embeddings, forming corresponding embeddings and targets for each encoder layer.

For the duration, we analyze the Whisper decoder embeddings, assuming that duration correlates with text-speech alignment learned through cross-attention. To generate target durations, we force-align the speech signal using WhisperX \cite{bain2023whisperxtimeaccuratespeechtranscription} and extract the duration of each word in the transcription. We then compute the mean decoder embeddings corresponding to each word, forming embeddings and targets for every decoder layer.

Finally, to capture non-linear dependencies, we train a random forest regression model for each encoder and decoder layer to predict prosodic targets from embeddings \cite{randomForests}. To estimate the $95\%$ confidence intervals of the predictions, we use $100$ bootstrap samples \cite{Efron1995AnIT}. Figure \ref{fig:encoder_decoder_analysis} summarizes model performance using Mean Average Error percentage (MAE\%) as the evaluation metric, where the MAE\% is a max-min normalization of each prosodic feature target. We find that deeper layers in both the encoder and decoder capture less  prosodic information in their embeddings.

\begin{table}[t!]
\centering
\caption{Word-level sentence stress detection of \method on \dataset by input layer embeddings. }
\label{tab:results-model-choice}
\resizebox{0.8\linewidth}{!}{
\begin{tabular}{l@{\hskip 7pt}c@{\hskip 7pt}c@{\hskip 9pt}c@{\hskip 9pt}c@{\hskip 7pt}c@{\hskip 9pt}c@{\hskip 9pt}}
\toprule
 \textbf{Model} & \textbf{Layer} &  \textbf{Prec} & \textbf{Rec} & \textbf{F1} \\
 \midrule
\multirow{4}{*}{\method} & 
3 & 0.781 & 0.624 & 0.693 \\
& 6 & 0.83 & 0.86 & 0.845 \\
& 9 & \textbf{0.912} & \textbf{0.906} & \textbf{0.909} \\
 & 12 & 0.902 & 0.868 & 0.884 \\
\bottomrule
\end{tabular}}
\end{table}

\subsection{Layer selection}
To further explore the relationship between prosodic information and stress detection, we investigate whether layers that capture prosody more effectively contribute to better stress detection. As shown in Table \ref{tab:results-model-choice}, using embeddings from earlier layers did not produce results comparable to those obtained from deeper encoder and decoder layers.

However, we observe that the $9$th layer outperforms the final $12$th layer. This finding suggests a potential trade-off between prosodic information in embeddings and the encoding of linguistic knowledge, where intermediate to final layers may offer the optimal balance for stress detection. This aligns with previous research \cite{pasad2021layer} on the wav2vec 2.0 speech representation transformer model \cite{baevski2020wav2vec}, which observed opposing trends between acoustic and semantic correlations across its layers.
\section{Results}
\label{sec:res}

We evaluate our model by comparing it to previously proposed approaches for sentence stress detection and established benchmarks. Our experiments across different datasets validate our model's ability to identify stressed words accurately while also demonstrating impressive generalization capabilities. We report performance using standard classification metrics: precision, recall and F$1$ score. In our settings, a word is considered stressed by the \method model if at least one of its tokens is marked as stressed. We use english-only model of Whisper small. All models were trained for $4$ epochs on each dataset, except for the model trained on \dataset for zero-shot evaluation, which was trained for two epochs.

\subsection{Datasets}
\label{sec:Benchmarks}
We consider \dataset and Aix-MARSEC \cite{Auran2004TheAP} for validation. In addition, we use Expresso \cite{nguyen2023expresso} and EmphAssess \cite{de2023emphassess} benchmarks for evaluation, described below.

\newpara{Aix-MARSEC} is a speech corpus containing over $5$ hours of $1980$s BBC radio recordings, featuring $53$ speakers across different speech styles \cite{Auran2004TheAP}. In accordance with previous work \cite{koreanarticle}, we define a word as stressed for the Aix-MARSEC ground-truth sentence-stress labels  if it contains the first syllable in Jassem’s narrow rhythm unit (NRU) notation \cite{jassem1952stress}. We split the recordings and their corresponding transcriptions into sentences and obtain $2400$ samples. We use $70\%$ of the samples for training and $30\%$ for testing, similar to previous work \cite{Lin2020JointDO}.

\newpara{Expresso.} An expressive speech dataset \cite{nguyen2023expresso} comprising $47$ hours of recordings from four speakers ($2$ female, $2$ male) across a diverse range of speaking styles, including both improvised and read speech. Following the filtering criteria in \cite{de2023emphassess}, we include only samples that contain at least one emphasized word. For fair comparison, we use the same test set configuration as \cite{de2023emphassess}, selecting the speakers with IDs \texttt{ex01} and \texttt{ex02}.

\newpara{EmphAssess.} A synthetically
generated dataset \cite{de2023emphassess} of speech sentences, each containing at least one stressed word. EmphAssess amounts to $3652$ samples, such that $913$ unique transcripts are rendered using $4$ distinct voices with varying stressed words. The voices used to synthesize all the transcripts, are the $4$ distinct Expresso voices, namely \texttt{ex01}, \texttt{ex02}, \texttt{ex03} and \texttt{ex04}. 

\subsection{Baselines}
\label{sec:Baseline}
We implement a simple BLSTM model with two layers, each containing $64$ hidden units, followed by a linear layer with two outputs for the stress detection task \cite{huang2015bidirectionallstmcrfmodelssequence}. Inspired by~\cite{Lin2020JointDO}, for each word in the transcription, we extract duration, mean energy and max pitch as audio features, in the corresponding speech segment. We train the baseline model using ground truth timestamps on each dataset. To further compare the baseline model with \method, which is alignment-free (i.e., it does not require timestamps for training or inference), we also evaluate the baseline models on a forced-aligned test set generated using the Montreal Forced Aligner (MFA) \cite{mcauliffe17_interspeech}. These models are referred to as \textit{Baseline (+GT alignment)} and \textit{Baseline (+MFA)}, respectively. We also compare \method against EmphaClass~\cite{de2023emphassess} and a Conditional Random Fields (CRF) model as in~\cite{koreanarticle}, and hierarchical BLSTM network as in \cite{Lin2020JointDO}.

\subsection{Results}

\newpara{Validation on \dataset.} We compare the performance of \method against the baseline outlined in Section \ref{sec:Baseline}, showing that \method achieves better results across all evaluation metrics. Furthermore, we show that in the common scenario where word-level timestamps are unavailable, force alignment may negatively impact the performance of models relying on their availability. Results are summarized in Table~\ref{tab:Expreeso-EmphaClass-AixMARSEC-tinystress-results}.

\newpara{Validation on Aix-MARSEC.} As summarized in Table \ref{tab:Expreeso-EmphaClass-AixMARSEC-tinystress-results}, \method achieves the best scores over all evaluation metrics compared to the BLSTM hierarchical network in \cite{Lin2020JointDO} and the CRF in \cite{koreanarticle}.
Additionally, we conduct analysis over the entire Aix-MARSEC dataset revealing that $\sim20$K of its $\sim50$K words are non-emphasized. However, the set of non-emphasized words contains only $\sim100$ unique words that frequently recur and are consistently labeled as non-emphasized. Therefore, emphasis detection over Aix-MARSEC dataset using NRU tagging is relatively straightforward. Consequently, it presents a scenario where words meaning plays a significant role in sentence-stress classification. This supports our hypothesis in Section \ref{sec:analysis} on the importance of language understanding captured by Whisper's decoder layers. 

\newpara{Evaluation on Expresso and EmphAssess.} For zero-shot evaluation, we use a \method variant trained solely on \dataset. Surprisingly, despite the distribution shift, \method achieves impressive zero-shot performance on both benchmarks, surpassing EmphaClass on Expresso. This while our method uses general-purpose synthetic data and EmphaClass was trained on a closed-source dataset with a distribution closely matching Expresso, inherently aligning it better with the evaluation data.

\begin{table}[t!]
    \caption{Word-level sentence-stress detection validation  on \dataset (denoted as TS-15K) and Aix-MARSEC datasets; evaluation on Expresso and EmphAssess. For Expresso and EmphAssess, we compare two variants of \method: (i) 0-shot and evaluation over all four speakers, and (ii) Training on speakers ex03, ex04, and evaluation over speakers ex01, ex02 (marked with *). Baseline variants are denoted "MFA" and "GT", as explained in section \ref{sec:Baseline}.}
    \label{tab:Expreeso-EmphaClass-AixMARSEC-tinystress-results}
    \centering
    \resizebox{\linewidth}{!}{
    \begin{tabular}{l@{\hskip 10pt}l@{\hskip 7pt}c@{\hskip 7pt}c@{\hskip 7pt}c@{\hskip 7pt}c}
        \toprule
        \textbf{Evaluated Dataset} & \textbf{Model} & \textbf{Prec} & \textbf{Rec} & \textbf{F1} \\
        \midrule
        \multirow{3}{*}{TS-15K}
        & Baseline (+GT alignment) & 0.862 & 0.853 & 0.858 \\
        & Baseline (+MFA) & 0.776 & 0.859 & 0.815 \\
 & \method & \textbf{0.912} & \textbf{0.906} & \textbf{0.909} \\
        \midrule
        \multirow{5}{*}{Aix-MARSEC}
        & CRF (N=5) \cite{koreanarticle} & 0.85 & 0.89 & 0.87\\
        & BLSTM \cite{Lin2020JointDO} & 0.88 & 0.92 & 0.90 \\
        & Baseline (+GT alignment) & 0.905 & 0.953 & 0.928 \\
        & Baseline (+MFA) & 0.904 & 0.958 & 0.930 \\
        & \method & \textbf{0.953} & \textbf{0.968} & \textbf{0.961} \\
        \midrule
        \midrule
        \multirow{3}{*}{Expresso} & EmphaClass & 0.569 & 0.769 & 0.654 \\
        & Baseline (+MFA) [0-shot] & 0.404 & 0.599 & 0.482 \\
        & \method [0-shot] & \textbf{0.573} & \textbf{0.863} & \textbf{0.689} \\
        \midrule
        \multirow{3}{*}{EmphAssess} & EmphaClass & 0.938 & 0.94 & 0.938 \\
        & Baseline (+MFA) [0-shot] & 0.566 & 0.609 & 0.587 \\
        & \method & \textbf{0.945*} & 0.942* & \textbf{0.943*} \\
        & \method [0-shot] & 0.672 & \textbf{0.98} & 0.797 \\
        \bottomrule
    \end{tabular}}
\end{table}

To assess \method in a similar setting on a task-oriented dataset, we train it on speaker IDs \texttt{ex03}, \texttt{ex04} of EmphAssess, which amount to $\sim1,500$ samples. 
Despite the limited data, \method surpasses EmphaClass when evaluated on the unseen speakers \texttt{ex01}, \texttt{ex02}. For Expresso, surpassing zero-shot performance requires significantly more samples and diversity of human-recorded speech with sentence stress annotations, which currently remain unavailable. Hence, we do not report results for \method using in-domain training data.
\section{Conclusion}
\label{sec:con}
In this work, we introduced \method, an extension to the Whisper model to enrich its transcriptions by marking sentence stress while preserving its core functionality. We also presented a fully automated pipeline for synthetically generating training data specific for sentence stress detection. Equipped with this automatic pipeline, we created \dataset, a synthetic dataset of $\sim$15 hours. When trained on \dataset, \method demonstrated strong performance and impressive zero-shot generalization across benchmarks, outperforming baselines that utilized training data with additional information or relying on closed-source, domain-targeted datasets. 

Analysis revealed that while deeper Whisper layers encode less prosodic information, they are more effective for stress detection, with intermediate layers striking the best balance. By eliminating the need for forced alignment or manual human annotations, \method provides a cleaner, more accessible and alignment-free approach to integrating sentence stress detection into \ac{ASR} systems. We hope our work will encourage further advancements in speech related tasks by enriching speech signal interpretation with semantically significant features.

\newpara{Acknowledgments.} This research work was supported by ISF grant 2049/22.

\bibliographystyle{IEEEtran}
\bibliography{mybib}

\end{document}